\documentclass[10pt,twocolumn,letterpaper]{article}

\usepackage{iccv}
\usepackage{times}
\usepackage{epsfig}
\usepackage{graphicx}
\usepackage{amsmath}
\usepackage{amssymb}
\usepackage{siunitx}
\usepackage{gensymb}

\usepackage{tabularx}
\usepackage{multirow}
\usepackage{booktabs}  
\usepackage{colortbl}  
\newcolumntype{C}{>{\centering\arraybackslash}X}

\usepackage{xcolor,colortbl}

\usepackage[pagebackref=true,breaklinks=true,letterpaper=true,colorlinks,bookmarks=false]{hyperref}

\iccvfinalcopy

\def\eg{\emph{e.g}\onedot} 
\def\ie{\emph{i.e}\onedot} 
 
\def\etc{\emph{etc}\onedot} 

\newcommand{\PAR}[1]{\vskip4pt \noindent {\bf #1~}}

\newcommand{\imagenet}{ILSVRC-2012}

\usepackage{amsmath}
\DeclareMathOperator*{\argmax}{arg\,max}

\usepackage{mathtools}

\DeclarePairedDelimiterX{\infdivx}[2]{(}{)}{%
  #1\;\delimsize\|\;#2%
}
\newcommand{\kl}{\mathrm{KL}\infdivx}

\begin{document}

\title{S$^\mathbf{4}$L: Self-Supervised Semi-Supervised Learning}

\author{Xiaohua Zhai\thanks{equal contribution}, Avital Oliver\footnotemark[1], Alexander Kolesnikov\footnotemark[1], Lucas Beyer\footnotemark[1]\\
Google Research, Brain Team\\
{\tt\small \{xzhai, avitalo, akolesnikov, lbeyer\}@google.com}
}

\maketitle

\begin{abstract}

This work tackles the problem of semi-supervised learning of image classifiers.
Our main insight is that the field of semi-supervised learning can benefit from the quickly advancing field of self-supervised visual representation learning.
Unifying these two approaches, we propose the framework of self-supervised semi-supervised learning ($S^4L$) and use it to derive two novel semi-supervised image classification methods.
We demonstrate the effectiveness of these methods in comparison to both carefully tuned baselines, and existing semi-supervised learning methods.
We then show that $S^4L$ and existing semi-supervised methods
can be jointly trained, yielding a new state-of-the-art result on semi-supervised \imagenet{} with 10\% of labels.
\end{abstract}


\section{Introduction}

Modern computer vision systems demonstrate outstanding performance on a variety of challenging computer vision benchmarks, such as image recognition~\cite{russakovsky2015imagenet}, object detection~\cite{lin2014microsoft}, semantic image segmentation~\cite{everingham2015pascal}, \etc.
Their success relies on the availability of a large amount of annotated data that is time-consuming and expensive to acquire.
Moreover, applicability of such systems is typically limited in scope defined by the dataset they were trained on.

Many real-world computer vision applications are concerned with visual categories that are not present in standard benchmark datasets, or with applications of dynamic nature where visual categories or their appearance may change over time.
Unfortunately, building large labeled datasets for all these scenarios is not practically feasible.
Therefore, it is an important research challenge to design a learning approach that can successfully learn to recognize new concepts by leveraging only a small amount of labeled examples.
The fact that humans quickly understand new concepts after seeing only a few (labeled) examples suggests that this goal is achievable in principle.

\begin{figure}[t]
  \begin{center}
    \includegraphics[width=1.0\linewidth]{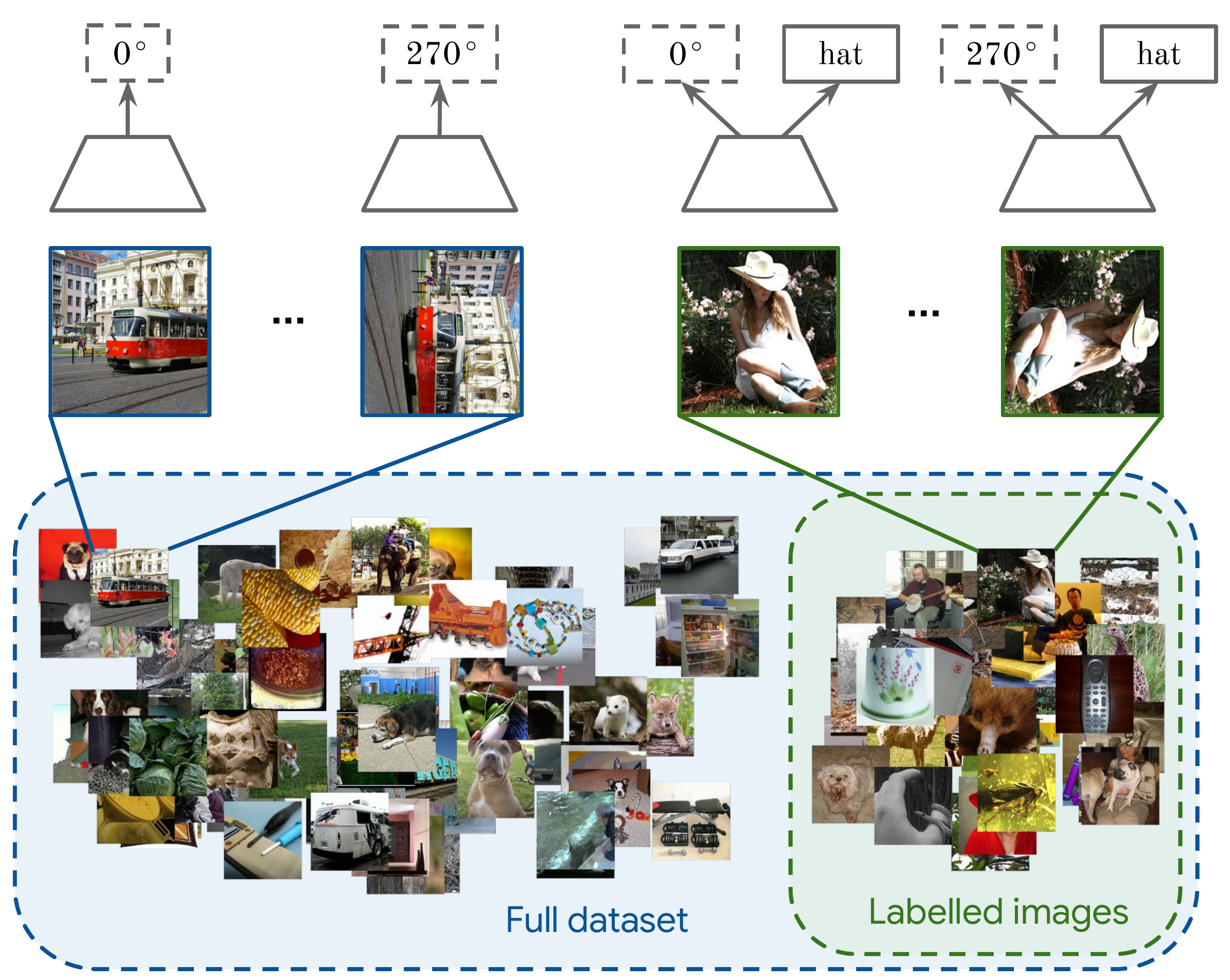}
  \end{center}
  \caption{
     A schematic illustration of one of the proposed self-supervised semi-supervised techniques: $S^4L$-Rotation.
     Our model makes use of both labeled and unlabled images.
     The first step is to create four input images for any image by rotating it by $0\degree$, $90\degree$, $180\degree$ and $270\degree$ (inspired by \cite{gidaris2018unsupervised}).
     Then, we train a single network that predicts which rotation was applied to all these images and, additionally, predicts semantic labels of annotated images.
     This conceptually simple technique is competitive with existing semi-supervised learning methods.
  }\label{fig:teaser}
\end{figure}

Notably, a large research effort is dedicated towards learning from unlabeled data that, in many realistic applications, is much less onerous to acquire than labeled data.
Within this effort, the field of self-supervised visual representation learning has recently demonstrated the most promising results~\cite{kolesnikov2019revisiting}.
Self-supervised learning techniques define \emph{pretext} tasks which can be formulated using only unlabeled data, but do require higher-level semantic understanding in order to be solved.
As a result, models trained for solving these pretext tasks learn representations that can be used for solving other downstream tasks of interest, such as image recognition.

Despite demonstrating encouraging results~\cite{kolesnikov2019revisiting}, purely self-supervised techniques learn visual representations that are significantly inferior to those delivered by fully-supervised techniques.
Thus, their practical applicability is limited and as of yet, self-supervision alone is insufficient.

We hypothesize that self-supervised learning techniques could dramatically benefit from a small amount of labeled examples.
By investigating various ways of doing so, we bridge self-supervised and semi-supervised learning, and propose a framework of semi-supervised losses arising from self-supervised learning targets. We call this framework self-supervised semi-supervised learning or, in short, $S^4L$.
The techniques derived in that way can be seen as new semi-supervised learning techniques for natural images.
Figure~\ref{fig:teaser} illustrates the idea of the proposed $S^4L$ techniques.  
We thus evaluate our models both in the semi-supervised setup, as well as in the transfer setup commonly used to evaluate self-supervised representations.
Moreover, we design strong baselines for benchmarking methods which learn using only \SI{10}{\percent} or \SI{1}{\percent} of the labels in \imagenet{}.

We further experimentally investigate whether our $S^4L$ methods could further benefit from regularizations proposed by the semi-supervised literature, and discover that they are complementary, \ie combining them leads to improved results.

Our main contributions can be summarized as follows:
\begin{itemize}
    \item We propose a new family of techniques for semi-supervised learning with natural images that leverage recent advances in self-supervised representation learning.
    \item We demonstrate that the proposed self-supervised semi-supervised ($S^4L$) techniques outperform carefully tuned baselines that are trained with no unlabeled data, and achieve performance competitive with previously proposed semi-supervised learning techniques.
    \item We further demonstrate that by combining our best $S^4L$ methods with existing semi-supervised techniques, we achieve new state-of-the-art performance on the semi-supervised ILSVRC-2012 benchmark.
\end{itemize}


\section{Related Work}

In this work we build on top of the current state-of-the-art in both fields of semi-supervised and self-supervised learning.
Therefore, in this section we review the most relevant developments in these fields.

\subsection{Semi-supervised Learning}

Semi-supervised learning describes a class of algorithms that seek to learn from both unlabeled and labeled samples, typically assumed to be sampled from the same or similar distributions.
Approaches differ on what information to gain from the structure of the unlabeled data.

Given the wide variety of semi-supervised learning techniques proposed in the literature, we refer to~\cite{semi_survey} for an extensive survey.
For more context, we focus on recent developments based on deep neural networks.

The standard protocol for evaluating semi-supervised learning algorithms works as such:
(1) Start with a standard labeled dataset;
(2) Keep only a portion of the labels (say, \SI{10}{\percent}) on that dataset;
(3) Treat the rest as unlabeled data.
While this approach may not reflect realistic settings for semi-supervised learning~\cite{oliver2018realistic}, it remains the standard evaluation protocol, which we follow it in this work.

Many of the initial results on semi-supervised learning with deep neural networks were based on generative models such as denoising autoencoders~\cite{ladder}, variational autoencoders~\cite{vae_ssl} and generative adversarial networks~\cite{gan_ssl1,gan_ssl2}.
More recently, a line of research showed improved results on standard baselines by adding \emph{consistency regularization losses} computed on unlabeled data.
These consistency regularization losses measure discrepancy between predictions made on perturbed unlabeled data points. Additional improvements have
been shown by smoothing predictions before measuring these perturbations.
Approaches of these kind include $\Pi$-Model~\cite{temporal_ensembling}, Temporal Ensembling~\cite{temporal_ensembling}, Mean Teacher~\cite{tarvainen2017mean} and Virtual Adversarial Training~\cite{vat}.
Recently, fast-SWA\cite{fast_swa} showed improved results by training with cyclic learning rates and measuring discrepancy with an ensemble of predictions from multiple checkpoints.
By minimizing consistency losses, these models implicitly push the decision boundary away from high-density parts of the unlabeled data.
This may explain their success on typical image classification datasets, where points in each clusters typically share the same class.

Two additional important approaches for semi-supervised learning, which  have shown success both in the context of deep neural networks and other types of models are Pseudo-Labeling~\cite{pseudo_label}, where one imputes approximate classes on unlabeled data by making predictions from a model trained only on labeled data, and conditional entropy minimization~\cite{entmin}, where all unlabeled examples are encouraged to make confident predictions on \emph{some} class.

Semi-supervised learning algorithms are typically~\cite{oliver2018realistic,vat,berthelot2019mixmatch, verma2019ict, fast_swa, liu2018deep} evaluated on small-scale datasets such as CIFAR-10~\cite{cifar10} and SVHN~\cite{svhn}.
We are aware of very few examples in the literature where semi-supervised learning algorithms are evaluated on larger, more challenging datasets such as ILSVRC-2012~\cite{russakovsky2015imagenet}.
To our knowledge, Mean Teacher~\cite{tarvainen2017mean} currently holds the state-of-the-art result on ILSVRC-2012 when using only \SI{10}{\percent} of the labels. Recent concurrent work \cite{xie2019unsupervised, henaff2019data} presents competitive results on ILSVRC-2012.


\subsection{Self-supervised Learning}

Self-supervised learning is a general learning framework that relies on surrogate (pretext) tasks that can be formulated using only unsupervised data.
A pretext task is designed in a way that solving it requires learning of a useful image representation.
Self-supervised techniques have a variety of applications in a broad range of computer vision topics~\cite{jang2018grasp2vec,sermanet2017time,ebert2018robustness,owens2018audio,sayed2018cross}.

In this paper we employ self-supervised learning techniques that are designed to learn useful visual representations from image databases.
These techniques achieve state-of-the-art performance among approaches that learn visual representations from unsupervised images only.
Below we provided a non-comprehensive summary of the most important developments in this direction.

Doersch et al. propose to train a CNN model that predicts relative location of two randomly sampled non-overlapping image patches~\cite{doersch2015unsupervised}.
Follow-up papers~\cite{noroozi2016unsupervised,noroozi2018boosting} generalize this idea for predicting a permutation of multiple randomly sampled and permuted patches.

Beside the above patch-based methods, there are self-supervised techniques that employ image-level losses.
Among those, in~\cite{zhang2016colorful} the authors propose to use grayscale image colorization as a pretext task.
Another example is a pretext task~\cite{gidaris2018unsupervised} that predicts an angle of the rotation transformation that was applied to an input image.

Some techniques go beyond solving surrogate classification tasks and enforce constraints on the representation space.
A prominent example is the \emph{exemplar} loss from~\cite{dosovitskiy2014exemplar} that encourages the model to learn a representation that is invariant to heavy image augmentations.
Another example is~\cite{noroozi2017representation}, that enforces additivity constraint on visual representation: the sum of representations of all image patches should be close to representation of the whole image.   
Finally, \cite{caron2018deep} proposes a learning procedure that alternates between clustering images in the representation space and learning a model that assigns images to their clusters.


\section{Methods}

In this section we present our self-supervised semi-supervised learning ($S^4L$) techniques.
We first provide a general description of our approach.
Afterwards, we introduce specific instantiations of our approach.

We focus on the semi-supervised image classification problem. Formally, we assume an (unknown) data generating joint distribution
$p(X, Y)$ over images and labels. The learning algorithm has access to a labeled training set $D_l$, which is sampled i.i.d. from $p(X, Y)$ and an unlabeled training set $D_u$, which is sampled i.i.d. from the marginal distribution $p(X)$.

The semi-supervised methods we consider in this paper have a learning objective of the following form:
\begin{equation}
    \min\limits_\theta \; \mathcal{L}_l(D_l, \theta) + 
                          w \mathcal{L}_u(D_u, \theta),
    \label{eq:ssss-objective}
\end{equation}
where $\mathcal{L}_l$ is a standard cross-entropy classification loss of all labeled images in the dataset, $\mathcal{L}_\textrm{u}$ is a loss defined on unsupervised images (we discuss its particular instances later in this section), $w$ is a non-negative scalar weight and $\theta$ is the parameters for model $f_\theta(\cdot)$.
Note that the learning objective can be extended to multiple unsupervised losses.

\subsection{Self-supervised Semi-supervised Learning}\label{sec:methods_$S^4L$}

We now describe our self-supervised semi-supervised learning techniques.
For simplicity, we present our approach in the context of multiclass image recognition, even though it can be easily generalized to other scenarios, such as dense image segmentation. 

It is important to note that in practice the objective~\ref{eq:ssss-objective} is optimized using a stochastic gradient descent (or a variant) that uses mini-batches of data to update the parameters $\theta$.
In this case the size of a supervised mini-batch $x_l, y_l \subset D_l$ and an unsupervised mini-batch $x_u \subset D_u$ can be arbitrary chosen.
In our experiments we always default to simplest possible option of using minibatches of equal sizes.

We also note that we can choose whether to include the minibatch $x_l$ into the self-supervised loss, i.e. apply $\mathcal{L}_\textrm{self}$ to the union of $x_u$ and $x_l$.
We experimentally study the effect of this choice in our experiments Section~\ref{sec:exp_$S^4L$}.

We demonstrate our framework on two prominent self-supervised techniques: predicting image rotation~\cite{gidaris2018unsupervised} and exemplar~\cite{dosovitskiy2014exemplar}. 
Note, that with our framework, more self-supervised losses can be explored in the future.

\PAR{$S^4L$-Rotation.}
The key idea of rotation self-supervision is to rotate an input image then predict which rotation degree was applied to these rotated images. The loss is defined as:
\begin{equation}
\mathcal{L}_{rot} = \frac{1}{|\mathcal{R}|} \sum_{r \in \mathcal{R}} \sum_{x \in D_u} \mathcal{L}(f_\theta(x^r), r) \label{eq:rotlossF}
\end{equation}
where $\mathcal{R}$ is the set of the $4$ rotation degrees $\{\SI{0}{\degree}, \SI{90}{\degree}, \SI{180}{\degree}, \SI{270}{\degree}\}$, $x^r$ is the image $x$ rotated by $r$, $f_\theta(\cdot)$ is the model with parameters $\theta$,$\mathcal{L}$ is the cross-entropy loss. This results in a 4-class classification problem.
We follow a recommendation from~\cite{gidaris2018unsupervised} and in a single optimization step we always apply and predict all four rotations for every image in a minibatch.

We also apply the self-supervised loss to the labeled images in each minibatch.
Since we process rotated supervised images in this case, we suggest to also apply a classification loss to these images.
This can be seen as an additional way to regularize a model in a regime when a small amount of labeled image are available.
We measure the effect of this choice later in Section~\ref{sec:exp_$S^4L$}.

\PAR{$S^4L$-Exemplar.}
The idea of exemplar self-supervision~\cite{dosovitskiy2014exemplar} is to learn a visual representation that is invariant to a wide range of image transformations.
Specifically, we use ``Inception'' cropping~\cite{szegedy2015going}, random horizontal mirroring, and HSV-space color randomization as in~\cite{dosovitskiy2014exemplar} to produce 8 different instances of each image in a minibatch.
Following~\cite{kolesnikov2019revisiting}, we implement $\mathcal{L}_u$ as the batch hard triplet loss~\cite{HermansBeyer2017Arxiv} with a soft margin.
This encourages transformation of the same image to have similar representations and, conversely, encourages transformations of different images to have diverse representations.

Similarly to the rotation self-supervision case, $\mathcal{L}_u$ is applied to all eight instances of each image.

\subsection{Semi-supervised Baselines}
In the following section, we compare $S^4L$ to several
leading semi-supervised learning algorithms that are not based on self-supervised objectives.
We now describe the approaches that we compare to.

Our proposed objective~\ref{eq:ssss-objective} is applicable for semi supervised learning methods as well, where the loss $\mathcal{L}_u$ is the standard semi supervised loss as described below.

\PAR{Virtual Adversarial Training} (VAT)~\cite{vat}: The idea is making the predicted labels robust around input data point against local perturbation.
It approximates the maximal change in predictions within an $\epsilon_{\text{vat}}$ vicinity of unlabeled data points, where $\epsilon_{\text{vat}}$ is a hyperparameter.
Concretely, the VAT loss for a model $f_\theta$ is:
\begin{equation}
  \mathcal{L}_{\text{vat}} = \frac{1}{|\mathcal{D}_u|} \sum_{x \in \mathcal{D}_u}\kl{f_\theta(x)}{f_\theta(x+\Delta x)},
\end{equation}
where
\begin{equation}
  \Delta x=\argmax_{\delta \text{ s.t. } |\delta|_2 = \epsilon} \kl{f_\theta(x)}{f_\theta(x+\delta)}.
\end{equation}
While computing $\Delta x$ directly is not tractable,
it can be efficiently approximated at the cost of an extra forward and backwards pass for every optimization step.~\cite{vat}.

\PAR{Conditional Entropy Minimization} (EntMin)~\cite{entmin}: This works under the assumption that unlabeled data indeed has one of the classes that we are training on, even when the particular class is not known during training.
It adds a loss for unlabeled data that, when minimized, encourages the model to make confident predictions on unlabeled data.
Specifically, the conditional entropy minimization loss for a model $f_\theta$ (treating $f_\theta$ as a conditional distribution of labels over images) is:
\begin{equation}
\mathcal{L}_{\text{entmin}} = \frac{1}{|\mathcal{D}_u|} \sum_{x \in \mathcal{D}_u} \sum_{y \in Y} -f_\theta(y|x) \log f_\theta(y|x) 
\end{equation}
Alone, the EntMin loss is not useful in the context of deep neural networks because the model can easily become extremely confident by increasing the weights of the last layer.
One way to resolve this is to encourage the model predictions to be locally-Lipschitz, which VAT does\cite{dirtt}.
Therefore, we only consider VAT and EntMin combined, not just EntMin alone, \ie $\mathcal{L}_u = w_{vat} \mathcal{L}_\textrm{vat} + w_\textrm{entmin} \mathcal{L}_\textrm{entmin}$.

\PAR{Pseudo-Label}~\cite{pseudo_label} is a simple approach: Train a model only on labeled data, then make predictions on unlabeled data. Then enlarge your training set
with the predicted classes of the unlabeled data points whose predictions are confident
past some threshold of confidence. Re-train your model with this enlarged labeled dataset.
While \cite{oliver2018realistic} shows that in a simple "two moons" dataset, psuedo-label fails to
learn a good model, in many real datasets this approach does show meaningful gains.

\section{\imagenet{} Experiments and Results}

In this section, we present the results of our main experiments.
We used the \imagenet{} dataset due to its widespread use in self-supervised learning methods, and to see how well semi-supervised methods scale.

Since the test set of \imagenet{} is not available, and numbers from the validation set are usually reported in the literature, we performed all hyperparameter selection for all models that we trained on a custom train/validation split of the public training set.
This custom split contains $1\,231\,121$ training and $50\,046$ validation images.
We then retrain the model using the best hyperparameters on the full training set ($1\,281\,167$ images), possibly with fewer labels, and report final results obtained on the public validation set ($50\,000$ images).

We follow standard practice~\cite{tarvainen2017mean,yunchen2016variational} and perform experiments where class-balanced labels are available for only \SI{10}{\percent} of the dataset.
Note that \SI{10}{\percent} of \imagenet{} still corresponds to roughly $128\,000$ labeled images, and that previous work uses the full (public) validation set for model selection.
While we use a custom validation set extracted from the training set, using such a large validation set does not correspond to a realistic scenario, as already discussed by~\cite{ladder,tarvainen2017mean,oliver2018realistic}.
We also want to cover more realistic cases in our evaluation.
We thus perform experiments on \SI{1}{\percent} of labeled examples (roughly $13\,000$ labeled images), while also using a validation set of only $5000$ images.
We analyze the impact of validation set size in Section~\ref{sec:valset}.

We always define epochs in terms of the available labeled data, \ie one epoch corresponds to one full pass through the labeled data, regardless of how many unlabeled examples have been seen.
We optimize our models using stochastic gradient descent with momentum on minibatches of size $256$ unless specified otherwise.
While we do tune the learning rate, we keep the momentum fixed at $0.9$ across all experiments.
Table~\ref{tbl:main} summarizes our main results.


\subsection{Plain Supervised Learning}\label{sec:exp_sup}

Whenever new methods are introduced, it is crucial to compare them against a solid baseline of existing methods.
The simplest baseline to which any semi-supervised learning method should be compared to, is training a plain supervised model on the available labeled data.

Oliver~\etal~\cite{oliver2018realistic} discovered that reported baselines trained on labeled examples alone are unfairly weak, perhaps given that there is not a strong community behind tuning those baselines. They provide strong supervised-only baselines for SVHN and CIFAR-10, and show that the gap shown by the use of unlabeled data is smaller than reported.

We observed the same in the case of \imagenet{}.
Thus, we aim to provide a strong  baseline for future research by performing a relatively large search over training hyperparameters for training a model on only \SI{10}{\percent} of \imagenet{}.
Specifically, we try weight-decay values in $\{1,3\} \cdot 10^{\{-2,-3,-4\}}$, learning rates in $\{0.3, 0.1, 0.03\}$, four different learning rate schedules spanning $30$ to $500$ epochs, and finally we explore various model architectures: ResNet50, ResNet34, ResNet18, in both ``regular'' (v1) and ``pre-activation'' (v2) variants, as well as half-, double-, triple-, and quadruple-width variants of these, testing the assumption that smaller or shallower models overfit less.

In total, we trained several thousand models on our custom training/validation split of the public training set of \imagenet{}.
In summary, it is crucial to tune both weight decay and training duration while, perhaps surprisingly, model architecture, depth, and width only have a small influence on the final results.
We thus use a standard, unmodified ResNet50v2 as model, trained with weight decay of $10^{-3}$ for $200$ epochs, using a standard learning rate of $0.1$, ramped up from $0$ for the first five epochs, and decayed by a factor of $10$ at epochs $140$, $160$, and $180$.
We train in total for $200$ epochs. 
The standard augmentation procedure of random cropping and horizontal flipping is used during training, and predictions are made using a single central crop keeping aspect ratio.

We perform a similar search when training our baseline on \SI{1}{\percent} of \imagenet{}, but additionally include two choices of data augmentation (whether or not to apply random color augmentation) and two minibatch sizes ($256$ and $1024$) in the hyperparameter search.
Perhaps somewhat surprisingly, the results here are similar, in that tuning the weight decay and training duration is crucial, but model architecture does not matter much.
Additionally, performing color augmentation becomes important.
Here too, we use a standard, unmodified ResNet50v2 as model, trained with weight decay of $10^{-2}$ for $1000$ epochs, using a learning rate of $0.01$\footnote{While the standard learning rate of $0.1$ worked equally well, learning curves seemed significantly less stable.}, ramped up from $0.0$ for the first ten epochs\footnote{This was likely not necessary, but kept for consistency.}, and decayed by a factor of $10$ at epochs $700$, $800$, and $900$.
We train in total for $1000$ epochs. 
A more detailed presentation of the results is provided in the supplementary material.

Using this only slightly altered training procedure, our baseline models achieve \SI{80.43}{\percent} top5 accuracy (\SI{56.35}{\percent} top1) on the public \imagenet{} validation set when trained on only \SI{10}{\percent} of the full training set.
Our \SI{1}{\percent} baseline achieves \SI{48.43}{\percent} top5 accuracy (\SI{25.39}{\percent} top1). 
These results form a solid baseline to compare to, considering that the same standard ResNet50v2 model achieves \SI{92.82}{\percent} top5 accuracy (\SI{75.89}{\percent} top1) on \SI{100}{\percent} of the labels.

For all further experiments, we reuse the best hyperparameters discovered here, except that we try two additional learning rates: $\{0.3, 0.1, 0.03\}$ for \SI{10}{\percent} and $\{0.03, 0.01, 0.003\}$ for \SI{1}{\percent}, and two additional weight decays: $\{10^{-4}, 3 \cdot 10^{-4}, 10^{-3}\}$ for \SI{10}{\percent} and $\{3 \cdot 10^{-3}, 10^{-2}, 3 \cdot 10^{-2}\}$ for \SI{1}{\percent}.
We also try two different weights $w_u$ for the additionally introduced loss $\mathcal{L}_u$: $w_u \in \{0.5, 1.0\}$.

\begin{table}[t]
  \setlength{\tabcolsep}{0pt}
  \setlength{\extrarowheight}{5pt}
  \renewcommand{\arraystretch}{0.75}
  \centering
  \caption{Top-5 accuracy [\%] obtained by individual methods when training them on \imagenet{} with a subset of labels. All methods use the same standard width ResNet50v2 architecture.}\label{tbl:main}
  \begin{tabularx}{\linewidth}{p{5.5cm}p{1pt}Cp{1pt}C}
    \toprule[1pt]
    \imagenet{} labels:        && \SI{10}{\percent} && \SI{1}{\percent} \\
    (i.e. images per class) && ($128$)           && ($13$)           \\
    \midrule

    Supervised Baseline (Section~\ref{sec:exp_sup}) && 80.43 && 48.43 \\

    \arrayrulecolor{lightgray}\midrule[0.25pt]\arrayrulecolor{black}

    Pseudolabels~\cite{pseudo_label}               && 82.41 && 51.56 \\
    VAT~\cite{vat}                        && 82.78 && 44.05 \\
    VAT + Entropy Minimization~\cite{entmin} && 83.39 && 46.96 \\

    \arrayrulecolor{lightgray}\midrule[0.25pt]\arrayrulecolor{black}

    Self-sup. Rotation~\cite{kolesnikov2019revisiting} + Linear    && 39.75 && 25.98 \\
    Self-sup. Exemplar~\cite{kolesnikov2019revisiting} + Linear    && 32.32 && 21.33 \\
    Self-sup. Rotation~\cite{kolesnikov2019revisiting} + Fine-tune && 78.53 && 45.11 \\
    Self-sup. Exemplar~\cite{kolesnikov2019revisiting} + Fine-tune && 81.01 && 44.90 \\

    \arrayrulecolor{lightgray}\midrule[0.25pt]\arrayrulecolor{black}

    $S^4L$-Rotation && 83.82 && 53.37 \\
    $S^4L$-Exemplar && 83.72 && 47.02 \\

    \bottomrule
  \end{tabularx}
\end{table}


\subsection{Semi-supervised Baselines}

We train semi-supervised baseline models using (1) Pseudo-Label, (2) VAT, and (3) VAT+EntMin.
To the best of our knowledge, we present the first evaluation of these techniques on \imagenet{}.

\PAR{Pseudo-Label}
Using the plain supervised learning models from Section~\ref{sec:exp_sup}, we assign pseudo-labels to the full dataset. 
Then, in a second step, we train a ResNet50v2 from scratch following standard practice, \ie with learning rate $0.1$, weight decay $10^{-4}$, and $100$ epochs on the full (pseudo-labeled) dataset.

We try both using all predictions as pseudo-labels, as well as using only those predictions with a confidence above $0.5$.
Both perform closely on our validation set, and we choose no filtering for the final model for simplicity.

Table~\ref{tbl:main} shows that a second step training with pseudo-labels consistently improves the results on both \SI{10}{\percent} and the \SI{1}{\percent} labels case. 
This motivates us to apply the idea to our best semi supervised model, which is discussed in Section~\ref{sec:exp_moam}.

\PAR{VAT} 
We first verify our VAT implementation on CIFAR-10.
With $4000$ labels, we are able to achieve \SI{86.41}{\percent} top-1 accuracy, which is in line with the \SI{86.13}{\percent} reported in~\cite{oliver2018realistic}. 

Besides the previously mentioned hyperparameters common to all methods, VAT needs tuning $\epsilon_\text{vat}$.
Since it corresponds to a distance in pixel space, we use a simple heuristic for defining a range of values to try for $\epsilon_\text{vat}$:
values should be lower than half the distance between neighbouring images in the dataset.
Based on this heuristic, we try values of $\epsilon_\text{vat} \in \{50, 50\cdot 2^{-1/3}, 50\cdot 2^{-2/3}, 25\}$ and found $\epsilon_\text{vat} \approx 40$ to work best.

\PAR{VAT+EntMin}
VAT is intended to be used together with an additional entropy minimization (EntMin) loss.
EntMin adds a single hyperparameter to our best VAT model:
the weight of the EntMin loss, for which we try $w_{\text{entmin}} \in \{0, 0.03, 0.1, 0.3, 1\}$, without re-tuning $\epsilon_\text{vat}$. 

The results of our best VAT and VAT+EntMin model are shown in Table~\ref{tbl:main}.
As can be seen, VAT performs well in the \SI{10}{\percent} case, and adding adding entropy minimization consistently improves its performance.
In Section~\ref{sec:exp_moam}, we further extend the co-training idea to include the self-supervised rotation loss. 


\subsection{Self-supervised Baselines}

Previous work has evaluated features learned via self-supervision on the unlabeled data in a ``semi-supervised'' way by either freezing the features and learning a linear classifier on top, or by using the self-supervised model as an initialization and fine-tuning, using a subset of the labels in both cases.
In order to compare our proposed way to do self-supervised semi-supervised learning to these common evaluations, we train a rotation and an exemplar model following the best practice from~\cite{kolesnikov2019revisiting} but with standard width (``$4\times$'' in~\cite{kolesnikov2019revisiting}).

Following our established protocol, we tune the weight decay and learning rate for the logistic regression, although interestingly the standard values from~\cite{he2016deep} of $10^{-4}$ weight decay and $0.1$ learning rate worked best.

The results of evaluating these models with both \SI{10}{\percent} and \SI{1}{\percent} are presented in Table~\ref{tbl:main} as ``Self-sup. + Linear'' and ``Self-sup. + Fine-tune''.
Note that while our results for the linear experiment are similar to those reported in~\cite{kolesnikov2019revisiting}, they are not directly.
This is due to 1) ours being evaluated on the public validation set, while they evaluated on a custom validation set, and 2) they used L-BFGS while we use SGD with standard augmentations.
Furthermore, fine-tuning approaches or slightly surpasses the supervised baseline.


\subsection{Self-supervised Semi-supervised Learning ($S^4L$)} \label{sec:exp_$S^4L$}

For training our full self-supervised semi-supervised models ($S^4L$), we follow the same protocol as for our semi-supervised baselines, \ie we use the best settings of the plain supervised baseline and only tune the learning rate, weight decay, and weight of the newly introduced loss.
We found that for both $S^4L$-Rotation and $S^4L$-Exemplar, the self-supervised loss weight $w=1$ worked best (though not by much) and the optimal weight decay and learning rate were the same as for the supervised baseline.

As described in Section~\ref{sec:methods_$S^4L$}, we apply the self-supervised loss on both labeled and unlabeled images.
Furthermore, both Rotation and Exemplar self-supervision generate augmented copies of each image, and we do apply the supervision loss on all copies of the labeled images.
We performed one case study on $S^4L$-Rotation in order to investigate this choice, and found that whether or not the self-supervision loss $\mathcal{L}_\textrm{self}$ is also applied on the labeled images does not have significant impact.
On the other hand, applying the supervision loss $\mathcal{L}_\textrm{sup}$ on the augmented images generated by self-supervision does indeed improve performance by almost \SI{1}{\percent}.
Furthermore, this allows to use multiple transformed copies of an image at inference-time (\eg four rotations) and take the average of their predictions.
While this 4-rot prediction is \SIrange{1}{2}{\percent} more accurate, the results we report do \emph{not} make use of this in order to keep comparison fair.

The results shown in Table~\ref{tbl:main} show that our proposed way of doing self-supervised semi-supervised learning is indeed effective for the two self-supervision methods we tried.
We hypothesize that such approaches can be designed for other self-supervision objectives.

We additionally verified that our proposed method is not sensitive to the random seed, nor the split of the dataset, see Appendix~\ref{sec:randomness} for details.

Finally, in order to explore the limits of our proposed models and match capacity of the architectures used in concurrent papers (e.g.~\cite{henaff2019data}), we train the $S^4L$-Rotation model with a more powerful architecture, such as ResNet152v2 $2\times$wider, and also use large computational budget to tune hyperparameters. In this case our model achieves even better results: \SI{86.41}{\percent} top-5 accuracy with \SI{10}{\percent} labels and \SI{57.50}{\percent} with \SI{1}{\percent} labels.


\section{Semi-supervised Learning is Complementary to $S^4L$} \label{sec:exp_moam}

\begin{table}[t]
  \setlength{\tabcolsep}{0pt}
  \setlength{\extrarowheight}{5pt}
  \renewcommand{\arraystretch}{0.75}
  \centering
  \caption{Comparing our MOAM to previous methods in the literature on \imagenet{} with \SI{10}{\percent} of the labels. Note that \emph{different models use different architectures}, larger than those in Table~\ref{tbl:main}.}\label{tbl:bigmodels}
  \begin{tabularx}{\linewidth}{p{5.0cm}p{0.2cm}Cp{0.2cm}Cp{0.2cm}C}
    \toprule[1pt]
    && labels && Top-5 && Top-1 \\
    \midrule

    MOAM full {\small(\emph{proposed})} && $10\%$ && \textbf{91.23} && \textbf{73.21} \\
    MOAM + pseudo label {\small(\emph{proposed})} && $10\%$ && 89.96 && 71.56 \\
    MOAM {\small(\emph{proposed})} && $10\%$ && 88.80 && 69.73 \\

    \arrayrulecolor{lightgray}\midrule[0.25pt]\arrayrulecolor{black}

    ResNet50v2 ($4\times$wider) && $10\%$ && 81.29 && 58.15 \\
    VAE + Bayesian SVM~\cite{yunchen2016variational} && $10\%$ && 64.76 && 48.41 \\
    Mean Teacher~\cite{tarvainen2017mean} && $10\%$ && 90.89 && - \\
    $^\dagger$UDA~\cite{xie2019unsupervised} && $10\%$ && 88.52\makebox[0pt]{\hspace{5pt}$^\dagger$} && 68.66\makebox[0pt]{\hspace{5pt}$^\dagger$} \\
    $^\dagger$CPCv2~\cite{henaff2019data} && $10\%$ && 84.88\makebox[0pt]{\hspace{5pt}$^\dagger$} && 64.03\makebox[0pt]{\hspace{5pt}$^\dagger$} \\

    \midrule
    \vspace{0.1mm}
    \emph{Training with all labels:} &&& \\
    
    \midrule

    ResNet50v2 ($4\times$wider) && $100\%$ && 94.10 && 78.57 \\
    MOAM {\small(\emph{proposed})} && $100\%$ && \textbf{94.97} &&  \textbf{80.17} \\
    $^\dagger$UDA~\cite{xie2019unsupervised} && $100\%$ && 94.45\makebox[0pt]{\hspace{5pt}$^\dagger$} && 79.04\makebox[0pt]{\hspace{5pt}$^\dagger$} \\
    $^\dagger$CPCv2~\cite{henaff2019data} && $100\%$ && 93.35\makebox[0pt]{\hspace{5pt}$^\dagger$} && - \\
    \bottomrule
    \multicolumn{7}{r}{\emph{\footnotesize{$^\dagger$ marks concurrent work.}}}
  \end{tabularx}
\end{table}

Since we found that different types of models perform similarly well, the natural next question is whether they are complementary, in which case a combination would lead to an even better model, or whether they all reach a common ``intrinsic'' performance plateau.

In this section, we thus describe our \emph{Mix Of All Models} (MOAM).
In short: in a first step, we combine $S^4L$-Rotation and VAT+EntMin to learn a $4\times$ wider~\cite{kolesnikov2019revisiting} model.
We then use this model in order to generate pseudo labels for a second training step, followed by a final fine-tuning step.
Results of the final model, as well as the models obtained in the two intermediate steps, are reported in Table~\ref{tbl:bigmodels} along with previous results reported in the literature.

\PAR{Step 1: Rotation+VAT+EntMin}
In the first step, our model jointly optimizes the $S^4L$-Rotation loss and the VAT and EntMin losses.
We iterated on hyperparameters for this setup in a less structured way than in our controlled experiments above (always on our custom validation set) and only mention the final values here.
Our model was trained with batch size $128$, learning rate $0.1$, weight decay $2 \cdot 10^{-4}$, training for $200$ epochs with linear learning rate rampup up to epoch $10$, then 10-fold decays at 100, 150, and 190 epochs.
We use inception crop augmentation as well as horizontal mirroring.
We used the following relative loss weights: $w_\text{sup} = 0.3$, $w_\text{rot}=0.7$, $w_\text{vat}=0.3$, $w_\text{entmin}=0.3$.
We tried a few heuristics for setting those weights automatically, but found that manually tuning them led to better performance.
We also applied Polyak averaging to the model parameters, choosing the decay factor such that parameters decay by \SI{50}{\percent} over each epoch.
Joint training of these losses consistently improve over the models with a single objective.
The model obtained after this first step achieves $88.80\%$ top-5 accuracy on the \imagenet{} dataset.

\PAR{Step 2: Retraining on Pseudo Labels}
Using the above model, we assign pseudo labels to the full dataset by averaging predictions across five crops and four rotations of each image%
\footnote{Generating pseudo-labels using 20 crops only slightly improved performance by \SI{0.25}{\percent}, but is cheap and simple.}.
We then train the same network again in the exact same way (\ie with all the losses) except for the following three differences:
(1) the network is initialized using the weights obtained in the first step
(2) every example has a label: the pseudo label
(3) due to this, an epoch now corresponds to the full dataset; we thus train for 18 epochs, decaying the learning rate after 6 and 12 epochs.

\PAR{Step 3: Fine-tuning the model}
Finally, we fine-tune the model obtained in the second step on the original \SI{10}{\percent} labels only.
This step is trained with weight decay $3 \cdot 10^{-3}$ and learning rate $5 \cdot 10^{-4}$ for 20 epochs, decaying the learning rate 10-fold every 5 epochs.

Remember that all hyper-parameters described here were selected on our custom validation set which is taken from the training set.
The final model ``MOAM (full)'' achieves \SI{91.23}{\percent} top-5 accuracy, which sets the new state-of-the-art. 


We conduct additional experiments and report performance
of MOAM (\ie only Step 1) with \SI{100}{\percent} labels in Table~\ref{tbl:bigmodels}.
Interestingly, MOAM achieves promising results even in the high-data regime with \SI{100}{\percent} labels, outperforming the fully supervised baseline: $+\SI{0.87}{\percent}$ for top-5 accuracy and $+\SI{1.6}{\percent}$ for top-1 accuracy.


\begin{figure}[t]
  \begin{center}
    \includegraphics[width=1.0\linewidth]{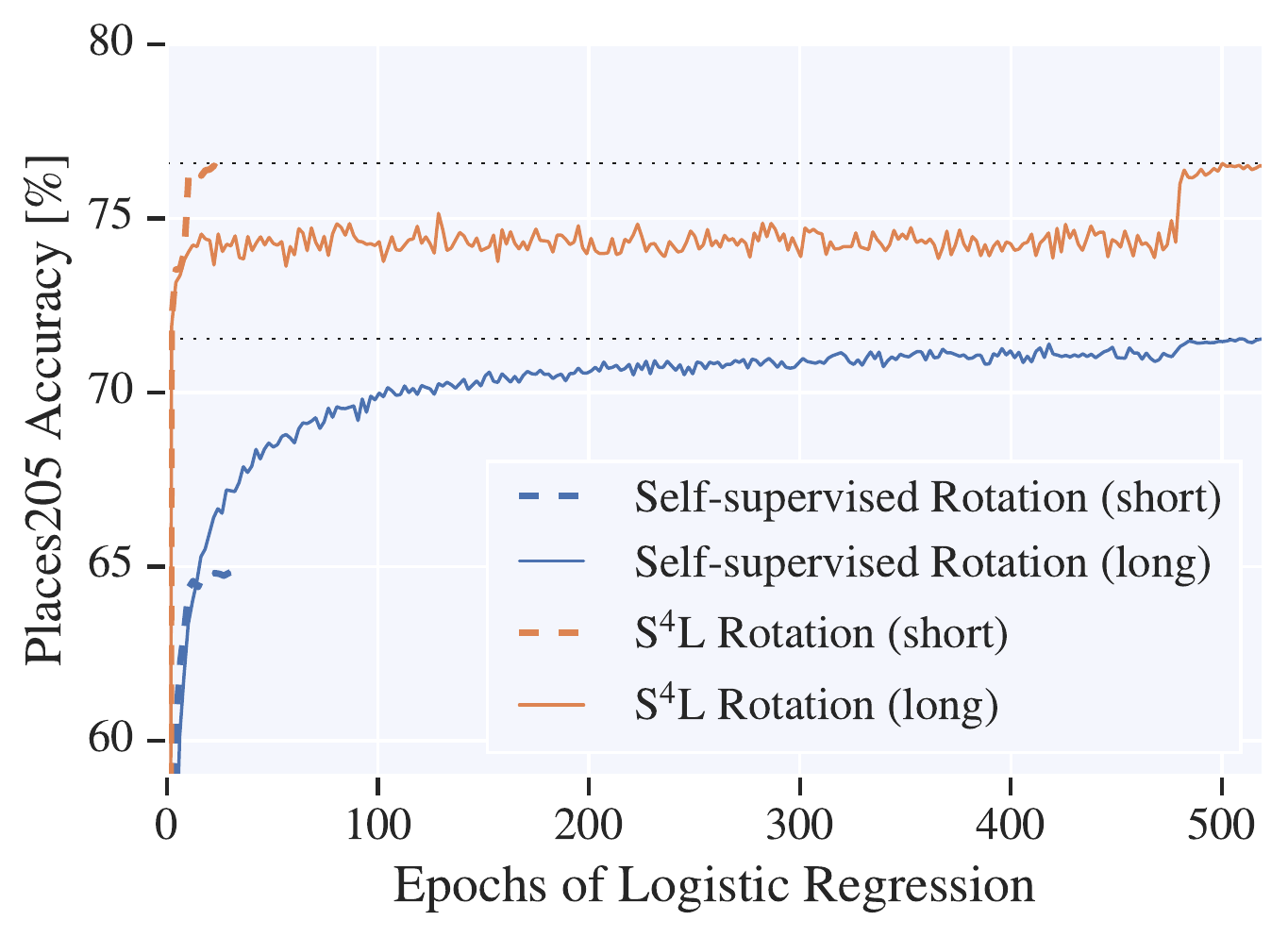}
  \end{center}
  \caption{Places205 learning curves of logistic regression on top of the features learned by pre-training a self-supervised versus $S^4L$-Rotation model on ILSVRC-2012.
  The significantly faster convergence (``long'' training schedule vs.\ ``short'' one) suggests that more easily separable features are learned.}\label{fig:transfer_curve}
\end{figure}

\section{Transfer of Learned Representations}

\begin{figure*}[t]
  \begin{center}
    \includegraphics[width=0.8\linewidth]{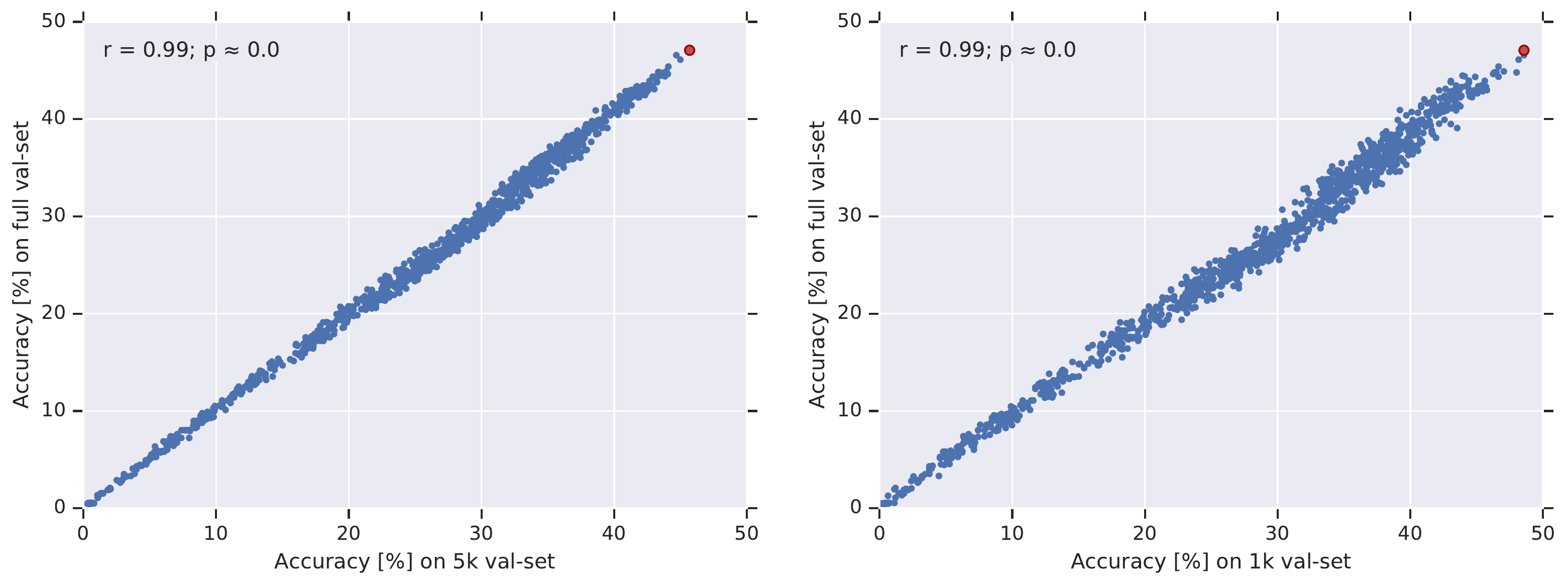}
  \end{center}
  \caption{Correlation between validation score on a (custom) validation set of $1000$, $5000$, and $50\,046$ images on \imagenet{}. Each point corresponds to a \emph{trained model} during a sweep for plain supervised baseline for the \SI{1}{\percent} labeled case.
  The best model according to the validation set of $1\,000$ is marked in red.
  As can be seen, evaluating our models even with only a single validation image per class is robust, and in particular
  selecting an optimal model with this validation set works as well as with the full validation set. 
  }\label{fig:valset}
\end{figure*}

Self-supervision methods are typically evaluated in terms of how generally useful their learned representation is.
This is done by treating the learned model as a fixed feature extractor, and training a linear logistic regression model on top the features it extracts on a different dataset, usually Places205~\cite{zhou2014learning}.
We perform such an evaluation on our $S^4L$ models in order to gain some insight into the generality of the learned features, and how they compare to those obtained by pure self-supervision.

We closely follow the protocol defined by~\cite{kolesnikov2019revisiting}.
The representation is extracted from the pre-logits layer.
We use stochastic gradient descent (SGD) with momentum for training these linear evaluation models with a minibatch size of 2048 and an initial learning rate of 0.1, warmed up in the first epoch.

While Kolesnikov~\etal~\cite{kolesnikov2019revisiting} show that a very long training schedule (520 epochs) is required for the linear model to converge using self-supervised representations, we observe dramatically different behaviour when evaluating our self-supervised semi-supervised representations. 
Figure~\ref{fig:transfer_curve} shows the accuracy curve of the plain self-supervised rotation method~\cite{kolesnikov2019revisiting} and our proposed $S^4L$-Rotation method trained on \SI{10}{\percent} of ILSVRC-2012.
As can be seen, the logistic regression is able to find a good separating hyperplane in very few epochs and then plateaus, whereas in the self-supervised case it struggles for a very long number of epochs.
This indicates that the addition of labeled data leads to much more separable representations, even across datasets.

We further investigate the gap between the representation learned by a good $S^4L$ model (MOAM) and a corresponding baseline trained on \SI{100}{\percent} of the labels (the baseline from Table~\ref{tbl:bigmodels}).
Surprisingly, we found that the representation learned by ``MOAM (full)'' transfers slightly better than the baseline, which used ten times more labelled data: \SI{83.3}{\percent} accuracy vs. \SI{83.1}{\percent} accuracy, respectively.
We provide full details of this experiment in the Supplementary Material.


\section{Is a Tiny Validation Set Enough?}\label{sec:valset}

Current standard practice in semi-supervised learning is to use a subset of the labels for training on a large dataset, but still perform model selection using scores obtained on the full validation set.\footnote{To make matters worse, in the case of \imagenet{}, this validation set is used both to select hyperparameters as well as to report final performance.
Remember that we avoid this by creating a custom validation set from part of the training set for all hyperparameter selections.}
But having a large labeled validation set at hand is at odds with the promised practicality of semi-supervised learning, which is all about having only few labeled examples.
This fact has been acknowledged by~\cite{ladder}, but has been mostly ignored in the semi-supervised literature. Oliver~\etal~\cite{oliver2018realistic} questions the viability of tuning with small validation sets by comparing the estimated model accuracy on small validation sets. They find that the variance of the estimated accuracy gap between two models
can be larger than the actual gap between those models, hinting that model selection
with small validation sets may not be viable. That said, they did not empirically evaluate whether it's possible to find the \emph{best} model with a small validation set, especially when choosing hyperparameters for a particular semi-supervised method.

We now describe our analysis of this important question. We look at the many models we trained for the plain supervised baseline on \SI{1}{\percent} of \imagenet{}.
For each model, we compute a validation score on a validation set of $1000$ labeled images (\ie one labeled image per class), $5000$ labeled images (\ie five labeled images per class), and compare these scores to those obtained on a ``full-size'' validation set of $50\,046$ labeled images.
The result is shown in Figure~\ref{fig:valset} and it is striking: there is a very strong correlation between performance on the tiny and the full validation set.
Especially, while in parts there is high variability, those hyperparameters which work best do so in either case. Most notably, the \emph{best model} tuned on a small validation set is also the best model tuned on a large validation set.
We thus conclude that for selecting hyperparameters of a model, a tiny validation set is enough.


\section{Discussion and Future Work}

In this paper, we have bridged the gap between self-supervision methods and semi-supervised learning by suggesting a framework ($S^4L$) which can be used to turn any self-supervision method into a semi-supervised learning algorithm.

We instantiated two such methods: $S^4L$-Rotation and $S^4L$-Exemplar and have shown that they perform competitively to methods from the semi-supervised literature on the challenging ILSVRC-2012 dataset.
We further showed that $S^4L$ methods are complementary to existing semi-supervision techniques, and MOAM, our proposed combination of those, leads to state-of-the-art performance.

While all of the methods we investigated show promising results for learning with \SI{10}{\percent} of the labels on ILSVRC-2012, the picture is much less clear when using only \SI{1}{\percent}.
It is possible that in this low data regime, when only $13$ labeled examples per class are available, the setting fades into the few-shot scenario, and a very different set of methods would be required for reaching much better performance.

Nevertheless, we hope that this work inspires other researchers in the field of self-supervision to consider extending their methods into semi-supervised methods using our $S^4L$ framework, as well as researchers in the field of semi-supervised learning to take inspiration from the vast amount of recently proposed self-supervision methods.

\PAR{Acknowledgements.} We thank the Google Brain Team in Z\"urich, and especially Sylvain Gelly for discussions.

{\small
\bibliographystyle{ieee}
\bibliography{egbib}
}

\clearpage
\appendix

\section{Detailed Results of the Supervised Baselines}

\begin{figure}[t]
  \begin{center}
    \includegraphics[width=1.0\linewidth]{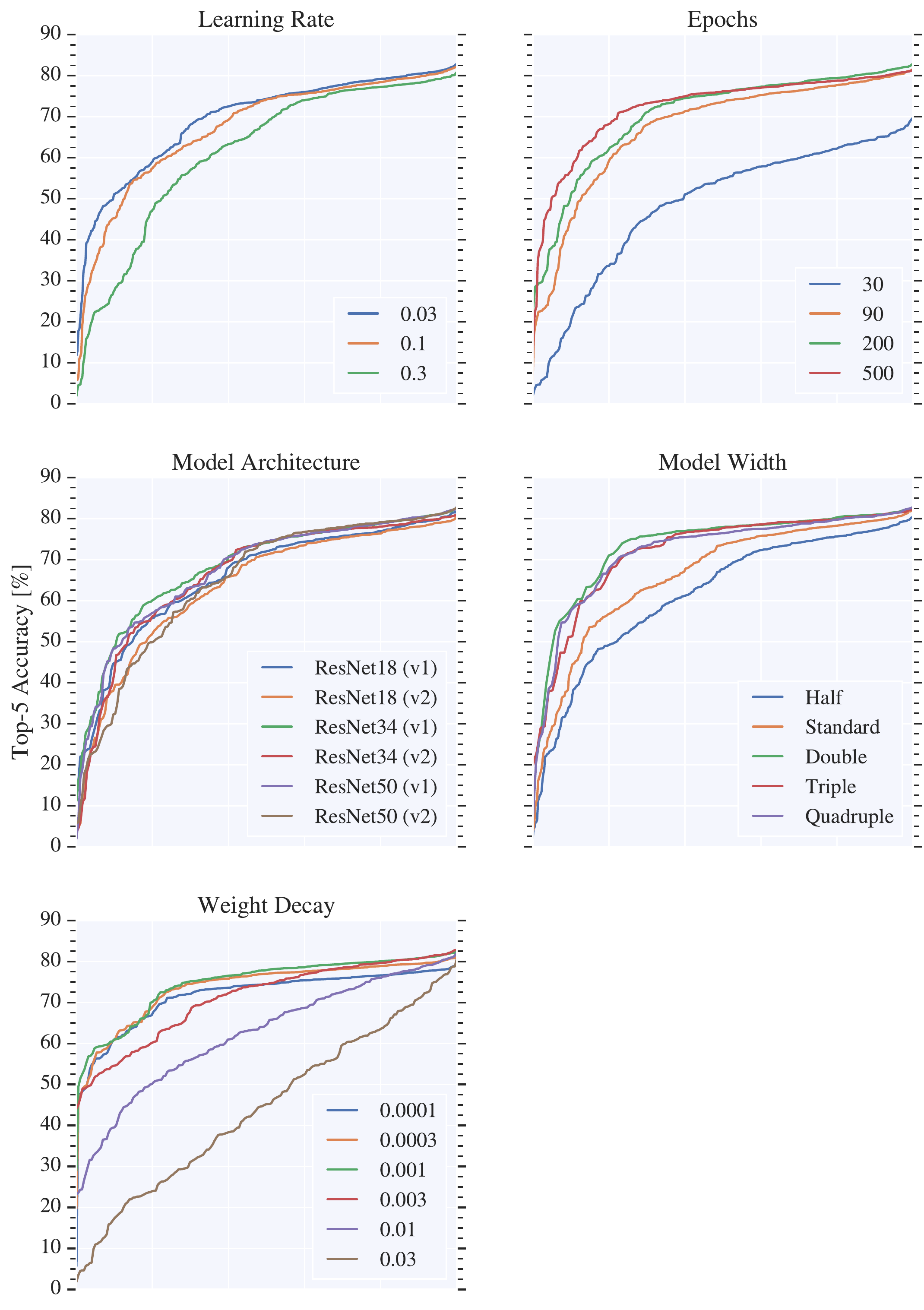}
  \end{center}
  \caption{The ``hypersweep curves'' for the supervised baseline trained on \SI{10}{\percent} of ILSVRC-2012. See text for details.}\label{fig:hypersweep_10p}
\end{figure}

\begin{figure}[t]
  \begin{center}
    \includegraphics[width=1.0\linewidth]{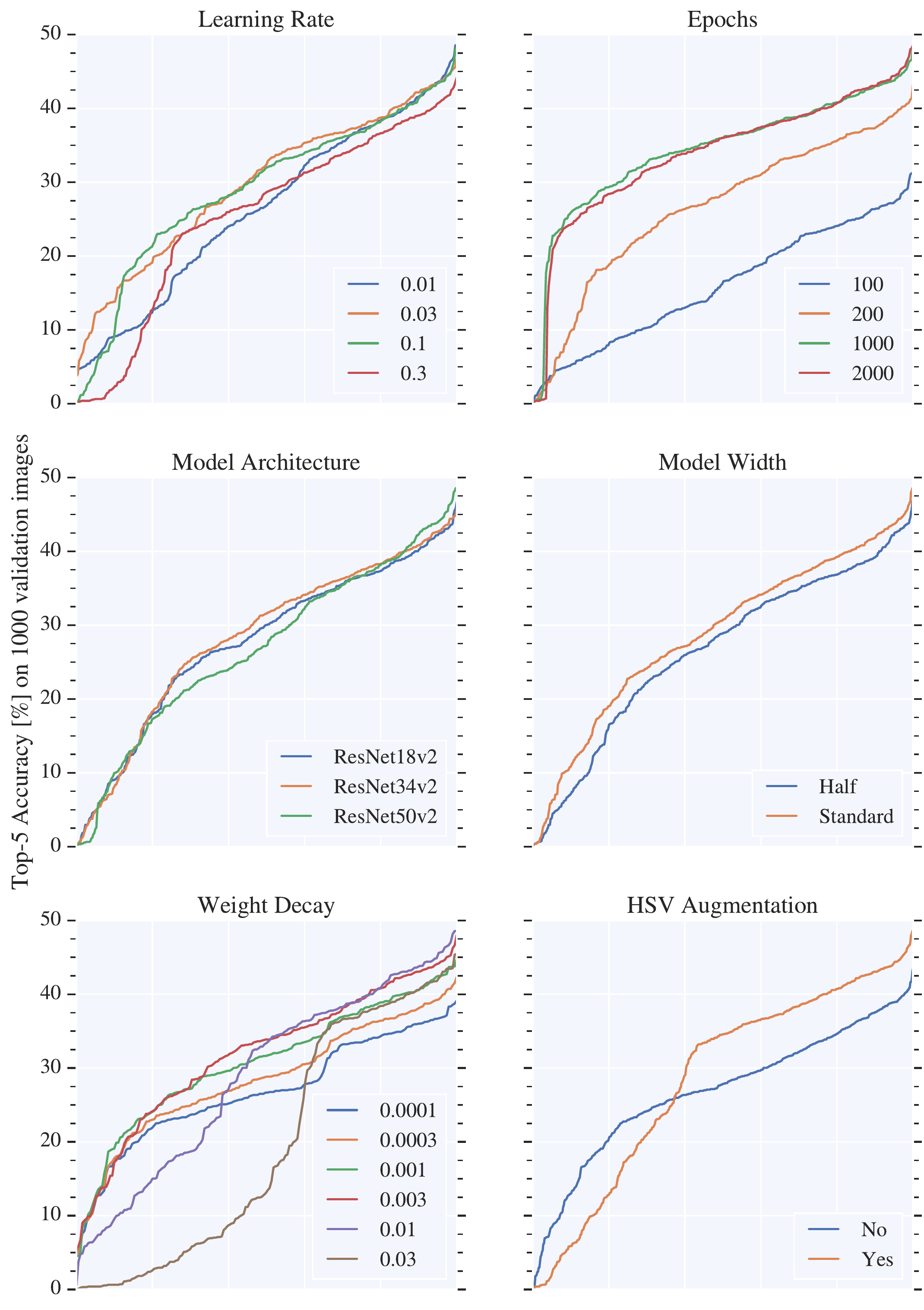}
  \end{center}
  \caption{The ``hypersweep curves'' for the supervised baseline trained on \SI{1}{\percent} of ILSVRC-2012. See text for details.}\label{fig:hypersweep_1p}
\end{figure}

Since we performed quite extensive hyperparameter search and trained many models in order to find a solid fully-supervised baseline on \SI{10}{\percent} and \SI{1}{\percent} of ILSVRC-2012, we believe that it is valuable to report the full results to the community, instead of just providing the final best model.

We present the results in the form of what we call ``hypersweep curves'' in Figures~\ref{fig:hypersweep_10p} and~\ref{fig:hypersweep_1p}.

Each plot shows a large collection of models -- \emph{each point on each plot is a fully trained model}. The curves are sorted by accuracy, allowing testing sensitivity to different hyperparameters, not only comparing the best model.
 
For each curve, we plot the accuracy of models where one of the hyperparameters is fixed.

Thus, by comparing curves, one can see:
\begin{enumerate}
    \item Which value of a hyperparameter performs best by looking at which curve's rightmost point is highest.
    \item How sensitive the model is to a hyperparameter \emph{in the best case} by looking at how far apart the curves are from eachother at their rightmost point.
    \item How robust a hyperparameter is \emph{on average} by looking at how similar the curves are overall.
    \item How independent a specific hyperparameter value is from all others by looking at the curve's shape, and whether curves cross-over (strong interplay) or not (strong independence).
\end{enumerate}

While the results shown in Figure~\ref{fig:hypersweep_10p} use the full (custom) validation set, those in Figure~\ref{fig:hypersweep_1p} were computed using the validation set of size $1000$, \ie with only one image per class.
As we have shown in Section~7, this is sufficient to determine the best hyperparameters, and we encourage the community to follow this more realistic protocol.

As can be seen, weight decay and number of training epochs are the two things which matter most when training using only a fraction of ILSVRC-2012.

Perhaps the most surprising finding is that, contrary to current folklore, \emph{reducing model capacity is detrimental} to performance on the smaller dataset.
Neither reducing depth, nor reducing width improve performance.
In fact, the deeper and wider models still outperform their shallower and thinner counterparts, even when using only \SI{1}{\percent} of the training data.
Even more so, the wider models are more robust to other hyperparameter's values as evidenced by their curves being significantly higher on the left end.
This is in line with recent findings suggesting wider models ease optimization~\cite{li2018visualizing,freeman2016topology,soltanolkotabi2019theoretical}.

Furthermore, when reducing the dataset size to \SI{1}{\percent}, we found that adding the same color augmentation as introduced by Exemplar is helpful.
We thereafter tried adding it to our best few models on \SI{10}{\percent}, but it did not help there.

Finally, while in the \SI{1}{\percent} case, learning-rate of $0.1$ and $0.01$ seem to perform equally well in the good cases (right hand side of curves), we manually inspected training curves and found that $0.1$ is significantly less robust, typically not learning anything before the first decay, and only catching up later on.

While we trained thousands of models in order to rigorously test multiple hypotheses (such as that of reducing model capacity), almost all boost in performance could have been achieved in just a few dozen trials with intuitively important hyperparameters (weight decay and epochs), which would take about a week on a modern four-GPU machine.

Overall, we hope that this thorough baseline investigation inspires the semi-supervised learning community to be more careful with baselines, as those that we found perform almost \SI{20}{\percent} absolute better than those previously reported in the literature.

\section{Randomness of $S^4L$}
\label{sec:randomness}

\begin{table}[h]
  \setlength{\tabcolsep}{0pt}
  \setlength{\extrarowheight}{5pt}
  \renewcommand{\arraystretch}{0.75}
  \centering
  \caption{$S^4L$ performance for 9 runs with random image subsets.
  Top-5 accuracies [\%] are reported as mean$\pm$standard deviation.}\label{tbl:subsets_and_seeds}
  \begin{tabularx}{\linewidth}{p{3.0cm}cp{3.0cm}cp{3.0cm}}
    \toprule[1pt]
    Method && 10\% ImageNet && 1\% ImageNet \\
    \midrule

    $S^4L$-Rotation && $83.91\pm0.13$ && $53.47\pm0.22$ \\

    $S^4L$-Exemplar && $83.76\pm0.06$ && $46.61\pm0.25$ \\

    \bottomrule
  \end{tabularx}
\end{table}

There are two factors of randomness of a semi supervised model: (1) labeled subset sampling, (2) run with different seeds. In order to estimate the randomness in the performance we train 9 models with random data subsets and random seeds for our proposed $S^4L$ method. Table~\ref{tbl:subsets_and_seeds} presents the detailed results. 
Overall, we observe that standard deviation is fairly small across both subsets and different runs and, therefore, our empirical evaluation provides robust comparison of various techniques.

\section{More Results in the Transfer Setup}

\begin{table}[t]
  \setlength{\tabcolsep}{0pt}
  \setlength{\extrarowheight}{5pt}
  \renewcommand{\arraystretch}{0.75}
  \centering
  \caption{Accuracy (in percent) obtained by various individual methods when transferring their representation to the Places205 dataset using linear models on frozen representations. All methods use the same plain ResNet50v2 base model, except for the ones marked by $^*$, which use a $4\times$ wider network. When it was necessary, a $^+$ marks longer transfer training of 520 epochs. The ``$\%$-labels'' column shows the percentage of ILSVRC-2012 labels that was used for training the model. }\label{tbl:transfer}
  \begin{tabularx}{\linewidth}{p{4.2cm}p{0.9pt}Cp{0.9pt}Cp{0.9pt}C}
    \toprule[1pt]
    Method        && \%-labels && top-5 && top-1 \\
    \midrule

    Supervised && 1 && 65.4 && 36.2 \\
    Supervised && 10 && 75.0 && 44.7 \\
    Supervised && 100 && 81.9 && 52.5 \\
    Supervised$^*$ && 100 && 83.1 && 53.7 \\
    
    \arrayrulecolor{lightgray}\midrule[0.25pt]\arrayrulecolor{black}

    SS Rotation$^+$~\cite{kolesnikov2019revisiting} && 0 && 71.4 && 41.7 \\    
    SS Exemplar$^+$~\cite{kolesnikov2019revisiting} && 0 && 69.0 && 39.8 \\
    
    \arrayrulecolor{lightgray}\midrule[0.25pt]\arrayrulecolor{black}

    Pseudolabels~\cite{pseudo_label}                && 1 && 71.6 && 41.8 \\
    VAT~\cite{vat}                         && 1 && 64.9 && 35.9 \\
    VAT + EntMin~\cite{entmin}                && 1 && 65.9 && 36.4 \\
    Pseudolabels~\cite{pseudo_label}                && 10 && 78.1 && 48.2 \\
    VAT~\cite{vat}                         && 10 && 76.4 && 45.8 \\
    VAT + EntMin~\cite{entmin}               && 10 && 76.4 && 46.2 \\
    
    \arrayrulecolor{lightgray}\midrule[0.25pt]\arrayrulecolor{black}
    
    SS Rotation~\cite{kolesnikov2019revisiting} + Fine-tune && 1 && 66.1 && 36.3 \\    
    SS Exemplar~\cite{kolesnikov2019revisiting} + Fine-tune && 1 && 60.0 && 31.1 \\
    SS Rotation~\cite{kolesnikov2019revisiting} + Fine-tune && 10 && 75.4 && 45.9 \\
    SS Exemplar~\cite{kolesnikov2019revisiting} + Fine-tune && 10 && 75.6 && 45.9 \\
        
    \arrayrulecolor{lightgray}\midrule[0.25pt]\arrayrulecolor{black}

    $S^4L$-Rotation && 1 && 67.3 && 38.0 \\
    $S^4L$-Exemplar && 1 && 61.2 && 32.2 \\
    $S^4L$-Rotation && 10 && 76.4 && 46.6 \\
    $S^4L$-Exemplar && 10 && 75.9 && 45.9 \\
    
    \arrayrulecolor{lightgray}\midrule[0.25pt]\arrayrulecolor{black}

    MOAM$^*$ full && 10 && 83.3 && 54.2 \\
    MOAM$^*$ + pseudo label && 10 && 83.3 && 54.2 \\
    MOAM$^*$ \begin{tabular}[t]{@{}c@{}}\hspace{-0.9em}\end{tabular} && 10 && 79.2 && 49.5 \\
    
    \bottomrule
  \end{tabularx}
\end{table}

In this section we present more results from the transfer evaluation task on Places205~\cite{zhou2014learning}. 
Table~\ref{tbl:transfer} shows the results for the models mentioned in our main paper. 
For each method, we select the best model and evaluate its transfer to Places205.

We follow the same setup as~\cite{kolesnikov2019revisiting} to train a linear models with SGD on top of frozen representations. 
The only difference is the training epochs, we train for 30 epochs in total with learning rate decayed at 10 and 20 epochs respectively. 
The learning rate is linearly ramped up for the first epoch. 
Kolesnikov et.al.~\cite{kolesnikov2019revisiting} train for 520 epochs with learning rate decays at 480 and 500 epochs. 
The schedule used in our paper is much shorter because of our finding that representation learned with labels are more separable and converges significantly faster.
(See in Section~6 of the main paper for details.)
To make fair comparison with the self-supervised models, results in Table~\ref{tbl:transfer} with $0\%$ labels are trained for 520 epochs to ensure their convergence.

From the plain supervised baselines, we observe that either more labels or wider networks lead to more transferable representations. 
Surprisingly, we found that pseudo labels outperforms the other two semi-supervised baselines in the transfer setup. 
On the $1\%$ labels evaluation setup, pseudo labels achieves the best result comparing to the other methods. 
With $10\%$ labels, $S^4L$ is comparable to the semi-supervised baselines, and our MOAM clearly outperforms all other models trained on $10\%$ of labels. 
More interestingly, the \emph{MOAM (full)} model on $10\%$ is slightly better than the $100\%$ supervised baseline with the same $4\times$ wider network. 
This indicates that learning a model with multiple losses may lead to representations that generalize better to unseen tasks.

\end{document}